# Trustworthy Clinical Decision Support Using Meta-Predicates and Domain-Specific Languages

*Michael Bouzinier[1,2,3], Sergey Trifonov[3], Michael Chumack[3], Eugenia Lvova[3,4], Dmitry Etin[3,4]*

Corresponding author: Michael Bouzinier

## Abstract

**Background**: Regulatory frameworks governing AI in healthcare, including the EU AI Act and FDA guidance on AI/ML-based medical devices, require clinical decision support systems to demonstrate not only accuracy but auditability. Existing formal languages for clinical logic validate syntactic and structural correctness but not whether decision rules use epistemologically appropriate evidence.

**Methods**: Drawing on design-by-contract principles, we introduce meta-predicates, that is predicates about predicates, as a mechanism for asserting epistemological constraints on clinical decision rules expressed in a domain-specific language (DSL). An epistemological type system classifies annotations along four dimensions: purpose, knowledge domain, biological scale, and method of acquisition, going beyond conventional data type checking. Meta-predicates, predicates about predicates, serve as design-by-contract constraints asserting which evidence types are permissible in any given rule. The framework is instantiated in AnFiSA, an open-source platform for genetic variant curation, and demonstrated using the Brigham Genomics Medicine protocol applied to 5.6 million variants from the Genome in a Bottle benchmark dataset.

**Results**: Decision trees used in variant interpretation can be reformulated as unate cascades, enabling complete per-variant audit trails that identify exactly which rule classified each variant and why. Meta-predicate validation catches epistemological errors before deployment, regardless of whether rules are human-written or AI-generated. Meta-predicates complement post-hoc explanation methods such as LIME and SHAP: where explanation reveals what evidence was used after the fact, meta-predicates constrain what evidence may be used before deployment, while preserving human readability of decision logic.

---

1 Harvard University, Cambridge, MA, USA

2 IDEXX Laboratories, Westbrook, ME, USA

3 Forome Association, Newton, MA, USA

4 Deggendorf Institute of Technology, Germany

**Conclusions**: We propose meta-predicate validation as a step toward addressing a key challenge in regulated healthcare AI: demonstrating not only that decisions are accurate but that they are made using appropriate evidence in ways that can be independently audited. While demonstrated in genomic variant classification, the approach generalises to any domain requiring auditable decision logic. The open-source implementation constitutes a proof of concept that we invite the community to extend and validate empirically across clinical domains.

# 1. Introduction

Healthcare systems increasingly rely on computational tools, including curated rule sets, annotation and data preparation pipelines, and machine learning components, to support clinical decisions from variant interpretation to treatment selection. As these systems grow more complex, particularly with artificial intelligence and machine learning components, their reliability, transparency, and accountability become critical. Regulatory frameworks in Europe (the AI Act, medical device documentation practices under the EU MDR, and emerging data governance initiatives such as the European Health Data Space) and in the United States (FDA guidance on AI/ML-based medical devices) now emphasize that clinical decision support systems must demonstrate not only accuracy but also auditability: organizations must be able to show how a decision was reached and why that process is trustworthy. In practice, this entails documenting what evidence was used, for what purpose, and why that evidentiary use is justified.

Our recent work on data preparation workflows [1] documented persistent challenges of provenance, traceability, and validation in health data pipelines. Current practice typically provides provenance as narrative descriptions tied to opaque code, which cannot support interpretive trust at scale. More generally, existing toolchains can document datasets, models, and code artifacts, but they still lack a practical mechanism for validating the decision logic itself, that is, whether a rule uses the right kind of evidence, at the right scale, and under an appropriate inferential pattern. While that earlier work addressed provenance in data transformation pipelines, this paper focuses on a related but distinct problem: how to formally validate the decision logic that operates on already prepared data.

In 2022 we introduced AnFiSA, an integrated development environment for genetic variant curation that incorporates a domain-specific language (DSL) for expressing classification rules [2]. That work focused primarily on system architecture, user interface, and deployment in clinical genomics workflows, and it described the mechanisms for automated verification of decision logic only informally, a gap that limits both the reproducibility of variant classification and the extension of the framework beyond genetics. Here we provide the formal foundations of that validation framework and generalize them beyond the specific AnFiSA implementation. Specifically, we formalize the DSL semantics together with an epistemological type system and meta-predicates that support prospective (design-time) validation of decision logic prior to deployment or reuse.

The aim is not to replace clinical judgment or regulatory review, but to provide a shared technical layer between those who author decision logic and those who must inspect, validate, or audit it. In this sense, the framework makes the evidentiary structure of decision rules explicit, traceable, and reproducible.

## 1.1 Contributions of This Work

This paper presents a formal specification and proof of concept implementation of a framework for validating clinical decision logic, using genetic variant classification as the primary use case. Our specific contributions are:

- **Epistemological type system and DSL**: We formalize the AnFiSA domain specific language (DSL) for variant classification and its epistemological type system, which classifies annotations by knowledge domain, scale, and method of acquisition.
- **Meta-predicates and validation semantics**: We introduce meta predicates, predicates about predicates, and define their validation semantics as machine checkable constraints on evidence used in decision rules. We implement a Python based validator that checks these constraints at design time.
- **Decision representations and traceability**: Building on the known equivalence between 1-decision lists, unate cascades, and nested canalyzing functions, we structure clinical decision rules as cascades and analyze how this representation enables fine grained per variant traceability and audit trails in our testbed system.
- **Prospective validation of human- and AI-generated logic**: We describe how the same validation mechanism applies prospectively to both human written and AI assisted decision logic, and we illustrate this in a genomics testbed rather than a live clinical deployment. We also discuss how the framework could constrain rules that are extracted from black box models.

While our concrete instantiation is in variant curation, we argue that the same principles extend to other domains that require validated decision logic, including clinical ethics compliance, bias detection, and regulatory adherence, and we sketch a general workflow for human–AI collaboration on decision rules in Section 5.

## 1.2 Motivating Example: Tracing a Variant Decision

To make the framework concrete before presenting its formal foundations, we show a complete decision trace for a single genetic variant. This example illustrates what validated, traceable clinical decision logic looks like in practice; the formal machinery that enables it is the subject of Sections 3 and 4.

The trace below follows the Brigham Genomics Medicine (BGM) protocol for identifying potentially damaging variants by the "Genetics First" approach [3]. The variant under consideration is a missense variant chr1:228287879 C>T in gene OBSCN, analyzed using the

Genome in a Bottle Ashkenazi Trio (NIST v. 4.2) [4]. The original NIST VCF contains 5,628,753 high quality variants. Tracing is performed by AnFiSA (backend v. 0.8.5, frontend v. 0.11.7).

The BGM protocol is described in [3] and in [2]. The rule used for illustration below is BGM "Red Button" described in detail in [2]. The rule selects 55 variants that have an elevated probability of causing adverse phenotypic effects.

AnFiSA implements BGM protocol as a sequential cascade of rules, each annotated with its epistemological properties: the type of knowledge used, the biological scale at which it operates, and the method by which it was obtained. The variant is tested against each rule in order; the first rule whose condition is met determines the outcome.

Table 1. Decision trace for variant chr1:228287879 C>T (OBSCN)

| Step | Test | Purpose / Knowledge Domain | Scale | Method | Action | Evaluated to: |
|---|---|---|---|---|---|---|
| 1 | Call Quality is poor | Provenance / Call Annotations | Variant | | **Reject** | False → Skip |
| 2 | Region is in masked repeats | Evidence / Human Genetics | Position | | **Reject** | False → Skip |
| 3 | Variant is called by a De-novo caller | Provenance / Call Annotations | Variant | Bioinformatics Inference | **Select** | False → Skip |
| 4 | Variant is called by a CNV caller | Provenance / Call Annotations | Variant | Bioinformatics Inference | **Select** | False → Skip |
| 5 | Variant is common (gnomAD AF >= 0.01) | Evidence / Population Genetics | Variant | Experimental, Other | **Reject** | False → Skip |
| 6 | gnomAD contains a homozygous record for this variant | Evidence / Population Genetics | Variant | Experimental, Other | **Reject** | False → Skip |
| 7 | gnomAD contains a hemizygous record for this variant | Evidence / Population Genetics | Variant | Experimental, Other | **Reject** | False → Skip |
| 8 | Variant is common for certain population (gnomAD_PopMax_AN >= 2000 and gnomAD_PopMax_AF >= .05) | Evidence / Population Genetics | Variant | Experimental, Other | **Reject** | False → Skip |
| 9 | Exclude non-coding variants, except those likely to alter splicing | Evidence / Functional Genetics | Variant | Bioinformatics Inference | **Reject** | False → Skip |
| 10 | Exclude known benign variants (Clinvar_Benign in {"Benign"} and Clinvar_stars in {"2", "3", "4"}) | Evidence / Human Genetics | Variant | Clinical Evidence | **Reject** | False → Skip |

| 11 | Exclude variants classified as benign by trusted submitters | Evidence / Human Genetics | Variant | Clinical Evidence | **Reject** | False → Skip |
| 12 | Select homozygous variants | Phenotype / Inheritance mode | Variant | | **Select** | False → Skip |
| 13 | Select x-linked variants | Phenotype / Inheritance mode | Variant | | **Select** | False → Skip |
| 14 | Select compound heterozygous variants | Phenotype / Inheritance mode | Variant in transcript | | **Select** | **True → Selected** |

*Some steps in the implemented protocol are condensed for presentation; for example, several individual quality checks are shown as a single step; the full current rule has 21 steps. Method column is the most important for evidence-related variables and sometimes not shown for variables serving other purposes.*

Several features of this trace merit attention. First, the decision path is linear: each rule is evaluated in sequence, and the variant's classification is determined by the first rule that fires (here, step 14). This structure is a unate cascade, discussed in Section 4, which makes it possible to identify exactly which rule classified any given variant and why.

Second, each step is annotated with its epistemological properties: purpose, knowledge domain, scale, and method. These annotations are not documentation; they are machine-checkable constraints (meta-predicates) that the validation engine verifies automatically. For instance, step 5 uses gnomAD allele frequency, which is classified as Population Genetics evidence at the variant scale. If a rule were modified to use a gene-scale metric while claiming variant-scale evidence, validation would fail before any variants were processed.

Third, the trace mixes different evidence types in a principled order: provenance checks first (steps 1–4), then population genetics filters (steps 5–8), then functional consequence filters (step 9), then clinical evidence (steps 10–11), and finally inheritance-based selection (steps 12–14). This ordering reflects clinical reasoning priorities and is enforced by the cascade structure.

The formal framework that enables this validation: the domain-specific language, epistemological type system, and meta-predicate semantics is presented in Section 3. The properties of the cascade representation and its role in traceability are discussed in Section 4.

## 1.3 From Trust Through Explanation to Trust Through Validation

Traditional approaches to trustworthy AI emphasize explainability: the ability to articulate why a system made a particular decision. This follows an intuitive path: explain → understand → trust. However, as models become more complex, full comprehension by domain experts becomes increasingly difficult, and narrative explanations can obscure rather than clarify the actual decision process.

We propose a complementary approach rooted in formal methods: formalize → validate → trust. Instead of focusing on post hoc explanations of individual decisions, we check whether decision logic written in our DSL satisfies explicit epistemological constraints via machine checked meta predicates, regardless of whether humans can follow every internal step.

Importantly, these approaches are not mutually exclusive. Our use of 1-decision lists maintains human readability (supporting intuitive trust) while meta-predicates enable automated validation (supporting formal trust).

## 1.4 Organization of this Paper

We have previously described a concrete instantiation of this framework in the AnFiSA system for variant curation [2]. In that work we present a realistic example and deployment details; here we focus on the formal underpinnings of the validation layer and its generalization beyond genomics.

We use genetic variant curation as our exemplar domain, because it provides clear evidence types, established clinical guidelines (ACMG/AMP), and well-defined classification outcomes. However, we believe that these principles generalize to any domain where decisions must satisfy certain trustability criteria:

- Based on specific types of evidence
- Traceable to their logical foundations
- Auditable for compliance with regulations or ethical standards
- Potentially generated or refined by AI systems

We discuss these broader applications in Section 5.3.

The remainder of this paper is organized as follows: Section 2 reviews related work in clinical decision support systems, formal methods in healthcare, and AI interpretability. Section 3 presents the complete specification of our DSL, including syntax, semantics, and the type system for biomedical evidence and introduces meta-predicates and their validation semantics. Section 4 describes the transformation from decision trees to unate cascades and the practical workflow for rule development and maintenance. Section 5 discusses implications for AI-generated decision logic, generalization to other domains, limitations, and future directions.

# 2. Related Work

Our framework sits at the intersection of clinical decision support, formal verification methods, and AI interpretability. We review relevant work in each area and position our contributions.

## 2.1 Clinical Decision Support, Variant Classification, and Formal Methods

Clinical decision support systems (CDSS) have a long history in healthcare, from early rule-based expert systems to modern AI-assisted tools. In genetic variant interpretation, the field has largely standardized around the ACMG/AMP guidelines [5], which classify variants as pathogenic, benign, or of uncertain significance based on multiple lines of evidence. However, these guidelines are expressed in natural language with terms like "very strong" and "moderate" evidence that require expert interpretation, leading to documented inter-laboratory disagreement [6], [7].

Computational tools for variant annotation, such as VEP [8], ANNOVAR [9], and annotation databases like ClinVar [10], provide the raw evidence but focus on evidence collection rather than evidence reasoning. Tools that attempt to automate ACMG guideline implementation, such as InterVar [11] and CharGer [12], hard-code interpretation rules in software rather than expressing them in a validatable language. Their logic cannot be systematically verified for epistemological appropriateness, nor can individual decisions be traced through the reasoning process in a uniform way.

Healthcare informatics has developed several formal languages for clinical logic. Table 2 compares the most prominent with our framework.

Table 2: Comparison of clinical decision support languages

| Feature | Arden Syntax | HL7 CQL | GLIF | FHIR Clinical Reasoning | This work |
|---|---|---|---|---|---|
| Syntactic validation | Yes | Yes | Yes | Yes | Yes |
| Data type checking | Yes | Yes | Partial | Yes | Yes |
| Epistemological type system | No | No | No | No | Yes |

| Feature | Arden Syntax | HL7 CQL | GLIF | FHIR Clinical Reasoning | This work |
|---|---|---|---|---|---|
| Meta-predicate constraints | No | No | No | No | Yes |
| Temporal reasoning | Limited | Yes | Yes | Yes | No |
| Guideline flowcharts | No | No | Yes | Partial | No |
| Per-decision traceability | No | Limited | Limited | Limited | Yes |

Arden Syntax [13] was among the earliest medical logic modules but focuses on triggering actions from simple conditions rather than validating evidence use. HL7 Clinical Quality Language (CQL) [14] provides strong typing and temporal reasoning but validates syntactic and structural correctness, not whether a decision uses the right kind of knowledge at the right scale. Guideline Interchange Format (GLIF) [15] represents guidelines as flowcharts with formal semantics and supports some consistency checking, but provides no mechanism for asserting constraints about evidence types at decision nodes. The FHIR Clinical Reasoning Module extends FHIR with decision support capabilities, often leveraging CQL, and shares its validation profile.

Though AnFiSA lacks some important functionality present in other systems, such as temporal reasoning and guideline flowcharts, it introduces a distinctive capability none of these approaches provide: meta-predicates that assert epistemological constraints on decision logic. This enables validation that checks not just whether code is well-formed, but whether it uses appropriate kinds of evidence—for example, ensuring that gene-level evidence is not used to make transcript-level decisions, or that clinical and population genetics evidence are not confused. Combined with cascade-structured decision logic, this provides per-decision traceability that goes beyond what existing languages offer.

## 2.2 Decision Representations: Trees, Lists, and Cascades

Decision trees are a classical representation for classification logic [16], [17], offering intuitive visualization and interpretability. However, complex decision trees can become difficult to trace, especially when decisions depend on deeply nested conditions.

Decision lists offer an alternative representation: an ordered sequence of if-then rules where the first matching rule determines the outcome. Rivest [18] proved that 1-decision lists—where each rule tests a single literal—have favorable learning properties, a result later expanded by Littlestone [19].

Independently, similar structures were studied in other fields under different names. Unate cascade functions were developed in the logic circuit design community for efficient circuit synthesis, while nested canalyzing functions were introduced by Kauffman et al. [20] to model gene regulatory networks in systems biology, where they exhibit stabilizing effects on network dynamics.

Jarrah et al. [21] proved that these apparently distinct concepts—unate cascade functions and nested canalyzing functions—are mathematically equivalent as classes of Boolean functions. This equivalence connects them to the 1-decision lists studied by Rivest, providing a unified mathematical framework across circuit design, machine learning, and biological modeling.

These representations align naturally with clinical reasoning patterns. Physicians often apply "inclusion and exclusion criteria" sequentially: first ruling out variants based on strong negative evidence, then progressively considering positive evidence. An approach to transformation of decision trees to unate cascades is discussed in Section 4 and exploits this structure to improve traceability: for any variant, we can show exactly which rule fired and why, rather than navigating a complex tree structure.

While the theoretical properties of unate cascades are well-established, their application to clinical decision support, and their combination with meta-predicate validation, is novel to our work.

In Section 4.1 we illustrate the transformation of decision tree logic to a unate cascade for genetic variant classification.

## 2.3 AI Interpretability and Program Synthesis

As machine learning models become more complex, extracting interpretable decision rules from black-box systems has become an active research area:

Decision tree extraction from neural networks was pioneered by Craven and Shavlik [22] and refined by subsequent work. Bastani et al.[23], [24] introduced techniques for building global explanations of complex models through enhanced decision tree construction, achieving both accuracy and interpretability. However, extracted trees are only as trustworthy as the validation framework that assesses them.

Program synthesis—automatically generating code from specifications—has shown promise for complex reasoning tasks. Recent work on the Arc Prize [25] demonstrates that deep learning-guided program synthesis can outperform test-time training and fine-tuning for abstract reasoning problems. This suggests a future where AI systems generate clinical decision logic rather than merely learning patterns from data.

Rule extraction and validation are complementary problems. While much AI interpretability work focuses on extracting human-readable rules from models, we focus on validating decision logic, whether human-written, AI-generated, or extracted from black-box systems. Our meta-predicates provide a formal framework for verifying that generated rules use evidence appropriately, regardless of their origin.

# 3. The AnFiSA Domain-Specific Language

We now present the formal specification of our DSL for expressing and validating clinical decision rules. We begin with an architectural overview, then detail the syntax, semantic model, and type system.

## 3.1 Architectural Overview

Our validation framework consists of four layers:

1. Data layer: Variant records with annotations from external tools (VEP, gnomAD, etc.)
2. Type system layer: Classification of annotations by epistemological properties
3. DSL layer: Scripts expressing decision logic with embedded meta-predicates
4. Validation layer: Automated checking that predicates conform to meta-predicate constraints

Figure 3.1 illustrates how these layers interact during variant classification.

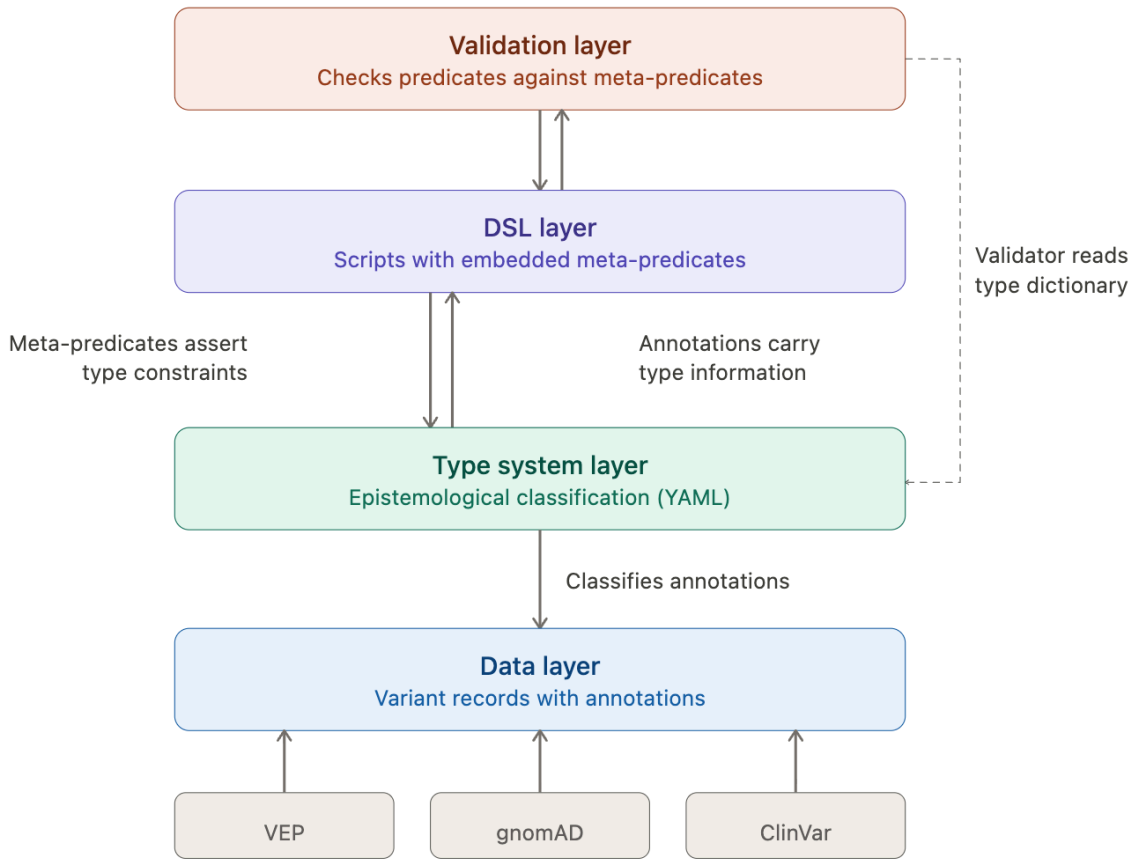

The key innovation is the bidirectional relationship between the type system and the DSL: annotations carry semantic type information (knowledge domain, scale, method), while meta-predicates in the DSL assert constraints on which types are permissible in decision logic. The validation layer mechanically verifies consistency.

## 3.2 Records and Variables

A record represents a single variant and its associated annotations. Structurally, a record is a flat collection of key-value pairs, where:

- Each **key** (also called an annotation) is a string from a predefined vocabulary
- Each **value** is a primitive type: string, integer, float, or boolean

For example:

```
{
  "pLI": 0.9,
  "QD": 4.2,
  "Most_Severe_Consequence": "stop_gained",
  "gnomAD_AF": 0.00001,
  "Gene": "CDH23"
}
```

A **variable** is a key-value pair within a record. We use "annotation" to refer to the key (e.g., pLI) and "variable" to refer to the key-value pair in context (e.g., **pLI = 0.9**).

Importantly, string values are conceptually **categorical** rather than free text: `Most_Severe_Consequence` can only take values from the Sequence Ontology [8], and `Gene` values come from HGNC nomenclature. Record values come from external annotation sources (VEP, gnomAD, etc.), which are responsible for vocabulary validation. Predicate expressions may reference arbitrary string constants syntactically, but these predicates will only evaluate to true when these constants match actual record values, providing natural validation through comparison semantics. When rules are constructed through the AnFiSA UI, valid values are presented via selection controls; the text editor allows free-form entry to support flexibility and extensibility.

## 3.3 DSL Syntax

The AnFiSA DSL is Python-based, leveraging Python's familiar syntax for logical expressions while adding specialized constructs for validation. A full language description including grammar is available in the AnFiSA GitHub repository (see Section 3.7); here we describe the key constructs informally.

### 3.3.1 Scripts and Pipelines

A **script** consists of a **pipeline** of statements followed by a final action:

```
<Script> ::= <Pipeline> <Action>
```

Each statement in the pipeline has three components:

1. Optional validation block (meta-predicates)
2. Predicate (logical expression)
3. Action (classification decision)

### 3.3.2 Statement Structure

A statement we used in Table 1 row 5 tests allele frequency (AF). Without validation this statement is:

```python
if (gnomAD_AF > 0.01):
    return False  # Exclude common variants
```

The same statement with validation:

```python
"""
@purpose(evidence)
@knowledge_domain("Population Genetics")
@scale(variant)
"""
if (gnomAD_AF > 0.01):
    return False
```

The validation block uses Python docstring syntax ("""...""") and contains one or more **meta-predicates**: assertions about the epistemological properties of variables used in the predicate.

### 3.3.3 Predicates

Predicates are Boolean expressions in standard Python syntax. They can use:

- Comparison operators: `<`, `>`, `<=`, `>=`, `==`, `!=`
- Logical operators: `and`, `or`, `not`
- Set membership: `in`, `not in`
- Parentheses for grouping

Variables are referenced by their annotation names (keys). For example:

```python
if (pLI > 0.9 and Most_Severe_Consequence in {
    'stop_gained', 'frameshift_variant'
} and gnomAD_AF < 0.001):
    return True
```

### 3.3.4 Actions

An action specifies the classification outcome:

- return True: include the variant (classify as potentially pathogenic)
- return False: exclude the variant (classify as benign or non-relevant)

### 3.3.5 Pipeline Execution Semantics

When a script is applied to a variant record, the pipeline executes sequentially:

1. Evaluate the first statement's predicate
2. If True, execute that statement's action and stop
3. If False, proceed to the next statement
4. If no predicate evaluates to True, execute the final action

This is first-match semantics: the first statement whose predicate is satisfied determines the outcome. This naturally implements a cascade or waterfall of decision rules.

### 3.3.6 Meta-Predicates

A meta-predicate has the form:

```
@<classification_dimension>(<classification_value>)
```

Where:

- `classification_dimension` ∈ {purpose, knowledge_domain, scale, method}
- `classification_value` is a valid value for that dimension (defined in Section 3.4)

Multiple meta-predicates can appear in a validation block.

## 3.4 Type System: Epistemological Classification of Annotations

Traditional programming type systems distinguish integers from strings, arrays from objects. Our type system goes further: it classifies variables by the epistemological nature of the evidence they represent.

### 3.4.1 Motivation

Consider two annotations: `gnomAD_AF` (allele frequency in a population database) and `PolyPhen` (computational prediction of protein function impact). Both are evidence about a variant's

pathogenicity, but they differ in several important ways:

- **Knowledge domain**: both gnomAD and PolyPhen are based on population genetics: gnomAD by directly measuring allele frequencies, PolyPhen by using evolutionary conservation across populations to infer functional constraint [26]. While PolyPhen is commonly assumed to be a functional predictor, its underlying mechanism is population-based, illustrating why domain classification requires explicit expert judgment rather than surface-level assumptions.
- **Scale**: gnomAD operates at the variant level; PolyPhen at the variant-in-transcript level
- **Method**: gnomAD is statistical genetics evidence; PolyPhen is bioinformatics inference

These distinctions matter clinically. ACMG guidelines specify which types of evidence support which conclusions. For example, population frequency evidence alone cannot establish pathogenicity (only refute it), while loss-of-function variants in dosage-sensitive genes require gene-level constraint metrics like pLI.

### 3.4.2 Classification Dimensions

Every annotation is classified along four dimensions:

1. **Purpose**
   - **phenotype**: Information about patient phenotype
   - **provenance**: Technical quality metrics (sequencing depth, mapping quality, etc.)
   - **evidence**: Data supporting pathogenicity assessment
2. **Knowledge Domain**
   - For Evidence:
     - **Human Genetics**: Observed disease associations
     - **Animal Genetics**: Model organism studies
     - **Population Genetics**: Allele frequency and constraint data
     - **Functional Genetics**: Consequence predictions and functional assays
     - **Epigenetics**: Gene expression, chromatin state
   - For Provenance:
     - **Call Annotations**: Technical variant calling metrics
   - For Phenotype:
     - **Phenotypic Data**: phenotypic data about the patient, proband or other study subject
     - **Inheritance Mode**: a mode of inheritance for a variant
3. **Scale**
   - **variant**: Properties of a specific variant
   - **position**: Properties of a genomic position
   - **transcript**: Properties within a specific transcript
   - **variant_in_transcript**: Variant effect on a specific transcript
   - **gene**: Properties of a gene
   - **window**: Properties of a genomic region

4. **Method** (primarily used for evidence)
   - **Clinical Evidence**: Observed in patients
   - **Statistical Genetics Evidence**: Population-level statistical inference
   - **Bioinformatics Inference**: Computational predictions
   - **Experimental, in Vivo**: Laboratory experiments in organisms
   - **Experimental, in Vitro**: Laboratory experiments in cell culture
   - **Experimental, Other**: Other types of experiments

### 3.4.3 Example Classifications

**Table 3: Example Annotation Classifications**

| Annotation | Purpose | Knowledge Domain | Scale | Method |
|---|---|---|---|---|
| gnomAD_AF | evidence | Population Genetics | variant | Statistical Genetics Evidence |
| pLI | evidence | Population Genetics | gene | Bioinformatics Inference |
| Most_Severe_Consequence | evidence | Functional Genetics | variant | Bioinformatics Inference |
| PolyPhen | evidence | Population Genetics | variant_in_transcript | Bioinformatics Inference |
| QD (Quality by Depth) | provenance | Call Annotations | variant | N/A |
| Mostly_Expressed_In | evidence | Epigenetics | gene | Experimental, in Vivo |

The complete classification dictionary contains over 100 annotations and is maintained as a YAML configuration file, allowing domain experts to extend or refine classifications without modifying code.

### 3.4.4 Type System Properties

Several properties ensure the type system is well-behaved:

1. **Completeness**: Every annotation used in decision logic must have a classification
2. **Unambiguity**: Each annotation has exactly one classification per dimension
3. **Orthogonality**: The four dimensions are independent (e.g., gene-scale evidence can

come from multiple knowledge domains)
4. **Extensibility**: New annotations and classification values can be added without breaking existing scripts

## 3.5 Validation Semantics

With the type system defined, we can now formalize validation:

**Definition (Valid Statement)**: *A statement is valid if and only if, for each meta-predicate in its validation block, there exists at least one variable in the predicate whose annotation's classification matches that meta-predicate.*

More formally, let:

- **M** = set of meta-predicates in the validation block
- **V** = set of variables used in the predicate
- **classify(v, d)** = the classification value of variable **v** along dimension **d**
- Then the statement is valid iff:

$$\forall m \in M : \exists v \in V : \text{classify}(v, \text{dimension}(m)) = \text{value}(m)$$

In the current implementation, the validation requires every declared meta-predicate to be satisfied by at least one variable. Though it might be a limitation in some cases, we believe this is an appropriate solution until the community adopts our proposal. Today, some rules legitimately mix evidence types in ways not yet fully systematized, so requiring every variable to satisfy a meta-predicate would generate too many false positives and discourage adoption. We discuss alternatives in section 5.5.1. We believe that best practices would avoid using differently typed variables in the same statement, if this recommendation is accepted, the validation can be made stricter.

### 3.5.1 Example: Successful Validation

```
"""
@purpose(provenance)
@knowledge_domain(Epigenetics)
@scale(gene)
@scale(variant)
"""
if ((0 < QD < 4) or Mostly_Expressed_In not in {"brain"}):
    return False
```

Validation check:

- `@purpose(provenance)`: QD has purpose=provenance ✓
- `@knowledge_domain(Epigenetics)`: Mostly_Expressed_In has domain=Epigenetics ✓
- `@scale(gene)`: Mostly_Expressed_In has scale=gene ✓
- `@scale(variant)`: QD has scale=variant ✓

All meta-predicates satisfied → statement is valid

### 3.5.2 Example: Validation Failure

```
"""
@purpose(evidence)
@knowledge_domain("Human Genetics")
@scale(variant)
"""
if (pLI < 0.9):
    return False
```

Validation check:

- `@purpose(evidence)`: pLI has purpose=evidence ✓
- `@knowledge_domain("Human Genetics")`: pLI has domain=Population Genetics ✗
- `@scale(variant)`: pLI has scale=gene ✗

Two meta-predicates unsatisfied → statement is **invalid**

The validation engine would report:

```
ValidationError in statement at line X:
  - No variable satisfies @knowledge_domain("Human Genetics")
  - No variable satisfies @scale(variant)
  Variables found: pLI (Population Genetics, gene)
```

## 3.6 Implementation Notes

The validation framework described in this paper is implemented as part of *AnFiSA* (Annotation and Filtration for Sequencing Analysis), an open-source platform for genetic variant curation described in detail in our earlier work [2]. Here we summarize the implementation status to clarify the boundary between what has been built and tested and what remains at the design stage.

The following components are fully implemented and have been used in operational variant curation workflows:

The DSL parser and execution engine process scripts written in the Python-based syntax described in Sections 3.2 and 3.3. The parser is using Python's ast module, which extracts variable references from predicates and validates syntactic correctness at load time, before any variants are processed. Additional parser components parse metapredicates within Python docstrings.

The epistemological type system (Section 3.4) is maintained as a YAML configuration file containing classifications for over 100 annotations along the four dimensions (purpose, knowledge domain, scale, method). This file is editable by domain experts without modifying code.

The meta-predicate validator checks each statement's predicate against its declared meta-predicates using the validation semantics defined in Section 3.5: every meta-predicate must be satisfied by at least one variable in the predicate. Validation runs at script load time and reports all violations with specific diagnostic messages.

The cascade execution engine evaluates scripts with first-match semantics and records complete decision traces (as illustrated in Table 1 and Figures 4.1 and 4.2). Per-step filtering statistics and interactive dashboards, illustrated in Figure 4.3, support population-level debugging.

The AnFiSA interface provides both a graphical rule editor (with selection controls that present valid annotation values) and a text editor for direct DSL authoring. The tracing facility allows users to enter a variant identifier and immediately see which rule classified it.

The following components are not yet implemented: exclusion meta-predicates (@exclude_domain, @exclude_variables), compliance meta-predicates (@preserves, @ensures), negation and disjunction over meta-predicate values, and the extensions to non-genomic domains sketched in Section 5.3. These require both formal specification and motivating use cases from the target domains.

## 3.7 Relevant GitHub Repositories

Forome GitHub (https://github.com/ForomePlatform) contains all relevant repositories. All source code is available at GitHub under Apache 2.0 license.

AnFiSA backend includes the cascade execution engine, the DSL Parser, the epistemological type system, and the meta-predicate validator and is located in: https://github.com/ForomePlatform/anfisa. The README file there guides through the steps to deploy the software.

The description of DSL and the epistemological type system can be found in: https://github.com/ForomePlatform/anfisa/blob/master/doc/specs/dsl.md

API Documentation is located in: https://github.com/ForomePlatform/documentation

Frontend including the rule editors and tracing facility is located in: https://github.com/ForomePlatform/Anfisa-React-Client

# 4. Transformation to Unate Cascades and Traceability

## 4.1 From Decision Trees to Unate Cascades

As discussed in Section 2.2, traditional decision trees for variant classification often have deeply nested structure. For example, a simplified pathogenicity assessment might be represented as:

```
if (gnomAD_AF > 0.01):
    EXCLUDE
else:
    if (Most_Severe_Consequence in LOF_SET):
        if (pLI > 0.9):
            INCLUDE
        else:
            if (ClinVar_Status == "Pathogenic"):
                INCLUDE
            else:
                EXCLUDE
    else:
        if (REVEL_score > 0.7):
            INCLUDE
        else:
            EXCLUDE
```

While this logic is sound, the nested structure complicates both traceability and validation. Following the decision path for a specific variant requires navigating multiple levels, modifying one branch can have unexpected effects on others, and verifying that each decision node uses epistemologically appropriate evidence requires examining variables scattered across the tree. The same logic can be reformulated as a sequential cascade:

```python
# Rule 1: Exclude common variants
if (gnomAD_AF > 0.01):
    return False

# Rule 2: Include high-confidence LOF in constrained genes
if (Most_Severe_Consequence in LOF_SET and pLI > 0.9):
    return True

# Rule 3: Include ClinVar pathogenic LOF
if (Most_Severe_Consequence in LOF_SET and ClinVar_Status == "Pathogenic"):
    return True

# Rule 4: Include damaging missense
if (REVEL_score > 0.7):
    return True

# Default: Exclude
return False
```

Rules are evaluated in order and the first match determines the outcome. This structure corresponds to what the literature calls unate cascade functions, nested canalyzing functions, and 1-decision lists, which Jarrah et al. [21] proved to be mathematically equivalent as classes of Boolean functions. The equivalence guarantees a single evaluation path per input and minimal evaluation cost.

Clinical decision rules frequently fit the cascade pattern naturally because they implement a hierarchy of evidence strength: strong negative evidence first (exclude common variants), then progressively weaker positive evidence. The BGM protocol trace in Table 1 (Section 1.2) illustrates this ordering. When rules cannot be expressed as cascades, the DSL supports nested conditionals, but traceability benefits are reduced.

## 4.2 Traceability, Auditing, and Visualization

The cascade structure enables complete per-variant audit trails, as shown in Table 1. For every variant processed, the system records which rules were evaluated, whether each condition was met, the actual variable values involved, and which rule ultimately fired. Figure 4.1 shows the AnFiSA cascade interface for the BGM protocol, displaying the waterfall of variant counts through each step alongside the DSL code and meta-predicate annotations for each rule. Figure 4.2 shows the per-variant trace for the example variant chr1:228287879 C>T, identifying the specific step at which it was classified. (Note: some steps in the implemented protocol are condensed into single rows in Table 1; for example, several individual quality checks are presented as one step.)

Beyond tracing individual variants, the cascade structure supports debugging at the population level. Each step partitions the remaining variants into those it catches and those that pass through, and the system provides detailed dashboards for each subset as shown on Figure 4.3. This is valuable for identifying edge cases: for example, examining variants excluded by a functional consequence filter might reveal that a particular frameshift variant was excluded because the frameshift occurs only in a non-canonical transcript and the variant is classified as benign in ClinVar — a finding that would be difficult to surface without per-step statistical breakdowns.

These traceability mechanisms support several audit scenarios. Clinical geneticists can verify that a variant was included or excluded for appropriate reasons. Compliance officers can check that decision logic conforms to guidelines such as ACMG criteria by examining meta-predicates against regulatory requirements. When classification accuracy is assessed, developers can identify which rules produce errors and refine them. And when rules are AI-generated or AI-refined, the audit trail combined with meta-predicate validation ensures that modifications remain epistemologically valid.

Figure 4.1 AnFiSA cascade interface for the BGM protocol (top part of the screen)

Figure 4.2 AnFiSA cascade interface for the BGM protocol (bottom part of the screen)

Figure 4.3 A dashboard describing variants selected at a certain step in the workflow

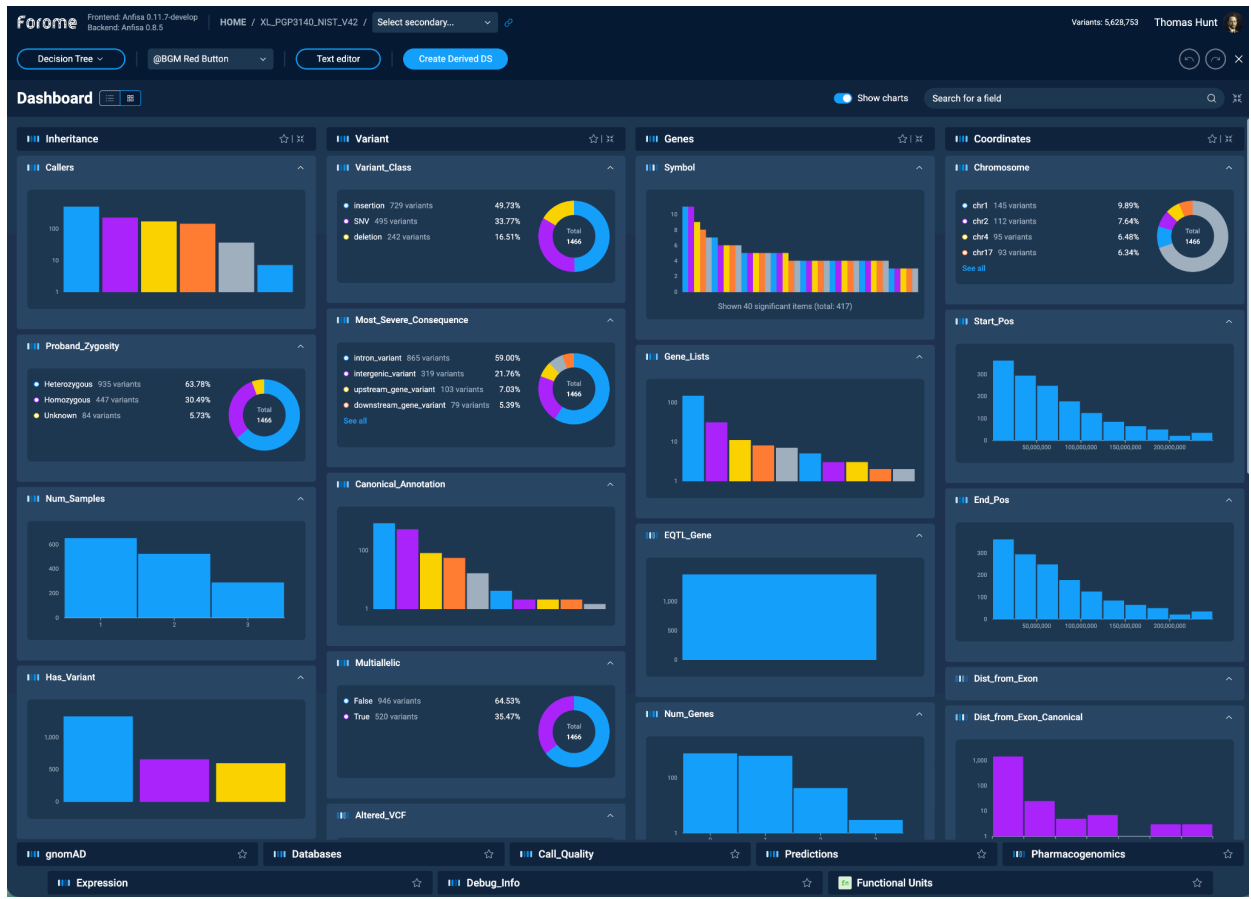

## 4.3 Practical Workflow for Rule Development

### 4.3.1 Development Process

The typical workflow for developing validated decision rules:

**Step 1: Domain expert specification**
A clinical geneticist writes rules in semi-formal language:

```
"Exclude variants with allele frequency > 1% in gnomAD,
 as these are too common to be pathogenic for rare diseases"
```

**Step 2: Initial implementation**
A bioinformatician translates to DSL:

```
if (gnomAD_AF > 0.01):
```

```
    return False
```

**Step 3: Meta-predicate annotation**
Together, they add epistemological constraints:

```
"""
@purpose(evidence)
@knowledge_domain("Population Genetics")
@scale(variant)
@method("Statistical Genetics Evidence")
"""
if (gnomAD_AF > 0.01):
    return False
```

**Step 4: Validation**
The system automatically verifies that gnomAD_AF satisfies all four meta-predicates. If validation fails, the team revisits the rule or the meta-predicates.

**Step 5: Testing**
The rule is tested on known variants (positive and negative controls) to ensure it behaves as intended.

**Step 6: Deployment**
Once validated, the rule is deployed in the classification pipeline.

### 4.3.2 Maintenance and Evolution

When rules need updating:

**Human modification**: If a geneticist modifies a threshold or adds a condition, meta-predicates are re-validated automatically. If new variables are introduced that violate meta-predicates, the system flags the error before deployment.

**AI-assisted refinement**: If an AI system suggests a rule modification to improve accuracy (e.g., based on retraining on new data), the modified rule must still satisfy the original meta-predicates. This prevents AI from introducing epistemologically invalid logic in pursuit of statistical optimization. This assumes that the AI system is constrained from modifying the meta-predicates themselves, a constraint that must be enforced at the deployment level.

**Versioning**: Each script version is tracked with its validation status, enabling audit trails that show when rules changed and why.

# 5. Discussion

## 5.1 Two Paths to Trust: Explanation and Validation

The challenge of building trustworthy AI systems has led to an emphasis on explainability: if we can articulate why a system made a decision, users will understand and trust it. This intuitive path: **explain → understand → trust** has driven much research in interpretable machine learning, from attention mechanisms to post-hoc explanation methods like Local Interpretable Model-agnostic Explanations (LIME) [27] and SHapley Additive exPlanations (SHAP) [28], [29].

However, this approach has inherent limitations. As models grow more complex, complete human comprehension becomes infeasible. A deep neural network with millions of parameters cannot be fully "understood" in the way a simple decision tree can. Moreover, explanations can be misleading: a system might provide a plausible-sounding justification that doesn't reflect its actual decision process, or cherry-pick features that correlate with decisions without revealing causal mechanisms.

We propose a complementary path grounded in formal methods: **formalize → validate → trust**. Rather than asking humans to understand every logical step, we express decision logic in a structured language with explicit epistemological constraints, then mechanically verify that these constraints are satisfied. Trust emerges not from comprehension but from demonstrated compliance with agreed-upon principles.

Importantly, these paths are not mutually exclusive—they address different trust requirements:

**Intuitive trust** (via explanation) answers: "Does this decision make sense to me as a domain expert?"

- Requires human-readable logic
- Supports clinical judgment and case-by-case review
- Essential for building confidence in individual decisions

**Formal trust** (via validation) answers: "Does this decision process conform to established standards?"

- Requires machine-verifiable constraints
- Supports regulatory compliance and systematic audit
- Essential for accountability at scale

The framework we propose supports both. The use of 1-decision lists, Python-based syntax, and tooling for traceability of the decision path for a given variant maintains readability for intuitive trust, while meta-predicates enable automated validation for formal trust. A clinical geneticist can read a rule and assess whether it "makes sense," while a compliance officer can verify that all rules use appropriate evidence types without manually reviewing every line of code.

We believe that this dual approach is particularly valuable in regulated healthcare environments, where organizations must demonstrate not only that their systems work, but that they work for the right reasons, using the right kinds of evidence, in ways that can be independently audited.

## 5.2 AI-Generated Decision Logic and Meta-Predicates

The emergence of capable AI systems raises a profound question: can machines generate trustworthy clinical decision logic?

### 5.2.1 From Black-Box Models to Validated Rules

Recent advances in model interpretability have focused on extracting decision rules from trained neural networks. Bastani et al. [23], [24] demonstrated that accurate decision trees can be extracted from complex models, providing global explanations of model behavior. However, extracted rules are only as trustworthy as the framework that validates them.

Our meta-predicates provide such a framework. Consider an AI system trained to classify variants that learns that replacing the following statement that uses VEP [8] annotations:

```
#Include Present in HGMD as "DM"
"""
@knowledge_domain("Human Genetics")
@scale("Variant")
@method("Clinical Evidence")
"""
if HGMD_Tags in {"DM"}:
    return True
```

With:

```
#Include variants with Damaging predictions
"""
@knowledge_domain("Human Genetics")
@scale("Variant")
@method("Clinical Evidence")
"""
if Polyphen_2_HVAR in {"P", "D"}:
    return True
```

Improves accuracy on a training set. Validation fails, flagging that the rule breaks all three assertions:

- Polyphen_2_HVAR is of scale "variant in transcript", not variant.
- Polyphen_2_HVAR comes from knowledge domain "Population Genetics" and obtained through method "Bioinformatics Inference"

A similar situation will be if AI replaces a block like:

```python
# 2.c. Include all potential LOF variants
# (stop-codon, frameshift, canonical splice site).
"""
@knowledge_domain("Functional Genetics")
@scale("Variant in Transcript")
@method("Bioinformatics Inference")
"""
if (Transcript_consequence in { 'transcript_ablation',
'splice_acceptor_variant', 'splice_donor_variant', 'stop_gained',
'frameshift_variant', 'stop_lost', 'start_lost' }):
    return True
```

With:

```python
# 2.c. Include all potential LOF variants
# (stop-codon, frameshift, canonical splice site).
"""
@knowledge_domain("Functional Genetics")
@scale("Variant in Transcript")
@method("Bioinformatics Inference")
"""
if (Canonical_Annotation in { 'transcript_ablation',
'splice_acceptor_variant', 'splice_donor_variant', 'stop_gained',
'frameshift_variant', 'stop_lost', 'start_lost' }):
    return True
```

This change will break the Scale meta-predicate, because the scale of the `Canonical_Annotation` variable is "Variant", not "Variant in Transcript".

To preserve validation, the AI must generate epistemologically valid logic, not just statistically accurate patterns.

## 5.2.2 Program Synthesis for Clinical Logic

Beyond extracting rules from trained models, recent work in program synthesis suggests AI can directly generate decision logic. The Arc Prize report [25] demonstrated that deep learning-guided program synthesis outperforms test-time training for complex reasoning tasks. This opens a provocative possibility: instead of training models to classify variants, we might train systems to generate classification rules that humans and machines can verify.

The advantages are significant:

- **Interpretability by construction**: Generated code is readable, not a black box
- **Formal verification**: Meta-predicates ensure generated rules use evidence appropriately
- **Iterative refinement**: Rules can be modified and re-validated as evidence evolves
- **Human-AI collaboration**: Experts can review, edit, and approve AI-generated rules

The key insight is that meta-predicates serve as **constraints on the program synthesis search space**. Rather than searching all possible Boolean functions, the AI searches only those that satisfy epistemological constraints. This both improves efficiency and ensures outputs are trustworthy.

## 5.2.3 Reasoning, Justification, and Confabulation

A philosophical consideration informs this pragmatic approach. A natural objection to AI-generated decision logic is that it lacks authentic clinical reasoning: rules extracted from a black-box model, or synthesised directly by an AI system, were not arrived at through genuine deliberation, and this limits their trustworthiness. Cognitive science research gives some reason, we suggest speculatively, to question how decisive this objection is. Mercier and Sperber [30] and Nisbett and Wilson [31] have argued, on different grounds, that human reasoning is itself often post-hoc justification rather than true logical deliberation; that experts frequently generate plausible accounts of decisions whose actual causal determinants they do not consciously access. If so, the distinction between authentic human reasoning and AI-generated justification may be less categorical than it appears.

It is worth noting, however, that this observation does not undermine the authority of meta-predicates. Meta-predicates are not claims about how a rule was reasoned into existence; they are prospective formal constraints that must be satisfied before a rule can be deployed, regardless of its origin. Whether the expert who assigned them was engaged in genuine deliberation or post-hoc rationalisation is beside the point: the constraints are evaluated mechanically against the type system, and a rule either satisfies them or it does not. The relevant criterion for trustworthy clinical decision support is not the authenticity of the reasoning process but the verifiability of its output.

## 5.3 Generalization Beyond Genetic Variant Classification

While we have demonstrated our framework using genetic variants, the underlying principles apply to any domain where decisions must be based on specific types of evidence, traceable to their logical foundations, and auditable for compliance with regulations or ethical standards. We briefly sketch three potential applications. Each would require extending the meta-predicate vocabulary with constructs not yet formally defined or implemented in our framework, notably exclusion constraints and domain-specific assertion types. The code fragments below are illustrative, not executable under the current system.

Consider clinical trial enrollment decisions. Regulations specify that certain evidence types must (or must not) inform eligibility:

```
"""
@purpose(evidence)
@domain("Clinical History")
@exclude_domain("Demographic")  # Cannot use race, ethnicity
"""
if (prior_cardiovascular_event and statin_intolerant):
    return ELIGIBLE
```

Another possible application is bias auditing that requires verifying that decisions don't depend on inappropriate factors:

```
"""
@purpose(evidence)
@domain("Clinical Indication")
@exclude_variables(["zip_code", "insurance_type"])
"""
if (HbA1c > 8.5 and diabetic_complications):
    return PRESCRIBE_INSULIN
```

The type system can flag when socioeconomic proxies influence clinical decisions, helping organizations identify and eliminate structural biases.

Our recent work on data provenance [1] demonstrated the need for validated transformation pipelines. Meta-predicates extend naturally to data processing:

```
"""
@transformation_type("Aggregation")
```

```
@preserves("Patient Count")
@ensures("No Re-identification Risk")
"""
def create_summary_table(patient_records):
    # Implementation with validation
```

This connects decision logic validation (this paper) to data transformation validation (our earlier work), providing comprehensive formal verification for health data pipelines.

Adapting our framework to new domains requires:

1. **Domain-specific type system**: Define epistemological dimensions relevant to the domain (analogous to our knowledge domain/scale/method classification)
2. **Vocabulary of constraints**: Establish what meta-predicates are meaningful (e.g., in finance: evidence type, time scale, risk category)
3. **Classification dictionary**: Annotate domain variables with their epistemological properties
4. **Validation rules**: Specify which combinations of evidence types are valid for which decisions

The DSL syntax and validation engine are domain-agnostic; only the type system requires domain expertise to define.

## 5.4 Advantages Over Existing Approaches

As summarized in Table 2, existing clinical decision support languages validate syntax and data types but not epistemological appropriateness of evidence. Our meta-predicates fill this gap by enabling prospective constraints on what kinds of evidence may appear in decision logic.

This is also complementary to model explanation methods such as LIME [27] and SHAP [28], [29], which reveal what features influenced a particular prediction retrospectively. Meta-predicates constrain what types of features are permitted to influence any prediction, prospectively and before deployment. Explanation methods do not prevent models from using inappropriate evidence; they only reveal that it happened.

Finally, our meta-predicates extend the design-by-contract tradition [32] from asserting properties about input/output behavior to asserting properties about the epistemological structure of predicates. A traditional assertion checks that an allele frequency falls between 0 and 1; a meta-predicate checks that the variable being tested is population genetics evidence at the variant scale. The former validates data; the latter validates reasoning.

## 5.5 Limitations and Future Directions

### 5.5.1 Current Limitations

The AnFiSA implementation of meta-predicates is a proof of concept, not a production-quality framework. We show that formal validation is possible but implementing it in real-world settings requires community participation and feedback. We address this in section 5.6. Even if the implementation was production-quality, significant limitations would still be taken into consideration.

**Scope of validation**: Meta-predicates validate epistemological appropriateness but not clinical correctness. A rule can satisfy all meta-predicates yet still embody poor clinical judgment (e.g., wrong threshold). Meta-predicates are necessary but not sufficient for trustworthy logic.

**Manual annotation**: Assigning meta-predicates requires domain expertise and is labor-intensive. While this ensures thoughtful validation, it limits scalability.

**Discrete decisions only**: The current framework handles Boolean classification but not probabilistic reasoning or continuous risk scores. Extensions to Bayesian decision logic would be valuable.

**Transformation limitations**: Not all decision trees can be converted to unate cascades without introducing additional variables. We provide guidelines but not a formal algorithm for determining transformability.

**Vocabulary validation**: String values in predicates are not validated against controlled vocabularies (Sequence Ontology, HGNC, etc.). This is delegated to data sources, but tighter integration would improve safety.

### 5.5.2 Future Work

There are several directions in which the proposed formal validation framework can evolve. Granted, some limitations must be addressed before evolution can provide value. However, progress in the following directions will also help to overcome many limitations.

**Probabilistic DSL extension**: Extending the framework to Bayesian networks and probabilistic graphical models would enable validation of uncertainty quantification in clinical decisions.

**Cross-domain type system**: Developing a unified epistemological classification that spans multiple clinical domains (genomics, imaging, EHR data) would enable integrated decision support across data types.

**Formal verification tools**: Integrating with proof assistants [33], [34] could enable formal verification that entire pipelines satisfy specified properties, going beyond individual rule validation.

**Standardization and regulatory frameworks**: Working with regulatory bodies (ISO, FDA, EMA) to establish meta-predicate validation as a component of AI/ML device approval would accelerate adoption and standardization.

## 5.6 Toward a Community Framework

We envision this work as the foundation for a community-driven framework where:

- Clinical specialties define domain-specific type systems
- Researchers contribute validated rule libraries
- AI systems generate and refine rules within epistemological constraints
- Regulators audit compliance through automated meta-predicate checking

The open-source AnFiSA platform enables this vision, but realization requires collaboration across clinical genetics, bioinformatics, AI research, and regulatory communities. We invite partners to experiment with the framework, extend it to new domains, and contribute to evolving standards for trustworthy clinical decision support.

For meta-predicates to be useful in real-world settings we need to conduct empirical validation studies. Trials are needed to compare error rates, audit efficiency, and clinical confidence between validated and unvalidated decision logic to quantify the practical value of the proposed framework. This is impossible without community adoption of the approach.

# 6. Conclusion

We have presented a formal framework for validating clinical decision rules through domain-specific languages and meta-predicates. By classifying evidence along epistemological dimensions—knowledge domain, scale, and method—our type system enables automated verification that decision logic uses appropriate evidence types. Meta-predicates serve as constraints on both human-written and AI-generated rules, ensuring formal compliance independent of authorship.

Transformation to unate cascades provides enhanced traceability, allowing per-variant audit trails that support both clinical review and regulatory compliance. While demonstrated using genetic variant classification, the framework's principles generalize to any domain requiring validated decision logic.

This work represents a proof of concept requiring community participation to realize its potential. We invite collaborators from clinical genetics, bioinformatics, AI research, and regulatory domains to experiment with the framework, contribute domain-specific type systems, and help establish standards for trustworthy clinical decision support in an era of AI-assisted healthcare.

## Availability of Code, Data, and Artefacts

All artefacts referenced in this paper are open source and available in the project GitHub repository:

https://github.com/ForomePlatform/anfisa

The VCF files used for illustration are available for download from the National Center for Biotechnology Information public FTP site:

ftp://ftp-trace.ncbi.nlm.nih.gov/giab/

These data were prepared and released by the National Institute of Standards and Technology (NIST) as part of the Genome in a Bottle (GIAB) project [4].

## Ethics Statement

This project used only publicly available data from the Personal Genome Project and the Genome in a Bottle project. All variant curation examples are purely illustrative and are not intended for clinical use or deployment in patient care.

## Author Contributions

**Michael Bouzinier**: Conceptualization; back-end implementation; use case selection and testing; manuscript writing.

**Dmitry Etin**: Conceptualization; manuscript writing; project management.

**Sergey Trifonov**: Proposed and conceptualized the use of unate cascades; led core back-end implementation.

**Michael Chumack**: Front-end implementation; unate cascades research.

**Eugenia Lvova**: Manuscript review; use case support; competitive technology research.

All authors reviewed and approved the final version of the manuscript.

# Acknowledgements

The epistemological type system for genetic variant curation was proposed by Prof. Shamil Sunyaev; without his continuous support this work could not have been completed. The authors are also grateful to colleagues at the Brigham Genomics Medicine (BGM) program, where much of the thinking about the type system took shape. This work received no external funding. The authors declare no competing interests.